\documentclass[oneside]{article}

\usepackage{PRIMEarxiv}

\usepackage[utf8]{inputenc} 
\usepackage[T1]{fontenc}    
\usepackage{hyperref}       
\usepackage{url}            
\usepackage{booktabs}       
\usepackage{amsfonts}       
\usepackage{nicefrac}       
\usepackage{microtype}      
\usepackage{lipsum}
\usepackage{fancyhdr}       
\usepackage{graphicx}       
\graphicspath{{media/}}     
\usepackage{multirow}
\usepackage{bigstrut}
\newcommand\tab[1][1cm]{\hspace*{#1}}

\usepackage{algorithm}
\usepackage{amsmath}
\usepackage{adjustbox}
\usepackage{multirow}
\usepackage{amssymb}
\usepackage{anyfontsize}
\usepackage{blindtext}
\usepackage{multirow}
\usepackage{amssymb}
\usepackage{pifont}
\usepackage[dvipsnames]{xcolor}
\usepackage[dvipsnames]{xcolor}
\usepackage[labelformat=simple]{subcaption}

\DeclareCaptionLabelFormat{subcaptionlabel}{\normalfont(\textbf{#2}\normalfont)}
\usepackage{algpseudocode} 
\usepackage{booktabs} 
\usepackage{color, colortbl}

\definecolor{Gray}{gray}{0.9}
\definecolor{LightCyan}{rgb}{0.88,1,1}

\title{Mitigating Degree Biases in Message Passing Mechanism by Utilizing Community Structures}

\author{
  Van Thuy Hoang and O-Joun Lee${}^{\dagger}$ \\
  Department of Artificial Intelligence, The Catholic University of Korea \\
  Bucheon-si, Gyeonggi-do 14662, Republic of Korea\\
  \texttt{\{hoangvanthuy90,ojlee\}@catholic.ac.kr} \\
  }

\begin{document}
\maketitle

\begin{abstract}

This study utilizes community structures to address node degree biases in message-passing (MP) via learnable graph augmentations and novel graph transformers.
Recent augmentation-based methods showed that MP neural networks often perform poorly on low-degree nodes, leading to degree biases due to a lack of messages reaching low-degree nodes.
Despite their success, most methods use heuristic or uniform random augmentations, which are non-differentiable and may not always generate valuable edges for learning representations.
In this paper, we propose Community-aware Graph Transformers, namely CGT, to learn degree-unbiased representations based on learnable augmentations and graph transformers by extracting within community structures. 
We first design a learnable graph augmentation to generate more within-community edges connecting low-degree nodes through edge perturbation.
Second, we propose an improved self-attention to learn underlying proximity and the roles of nodes within the community.
Third, we propose a self-supervised learning task that could learn the representations to preserve the global graph structure and regularize the graph augmentations.
Extensive experiments on various benchmark datasets showed CGT outperforms state-of-the-art baselines and significantly improves the node degree biases.
The source code is available at \url{https://github.com/NSLab-CUK/Community-aware-Graph-Transformer}.

\let\thefootnote\relax
\footnotetext{${}^{\dagger}$ Correspondence: \texttt{ojlee@catholic.ac.kr}; Tel.: +82-2-2164-5516} 
\end{abstract}


\section{Introduction}

The message-passing (MP) mechanism has been widely used in Graph Neural Networks (GNNs) and achieved great success in numerous domains \cite{s23084168}.
GNNs learn representations for each target node through recursively receiving and aggregating the messages from its neighbors.
Thus, these GNNs primarily rely on the message information exchanged between pairs of nodes to update representations.

\begin{figure}[tb]
\centering
\begin{subfigure}{0.47\textwidth}
    \includegraphics[width=\textwidth]{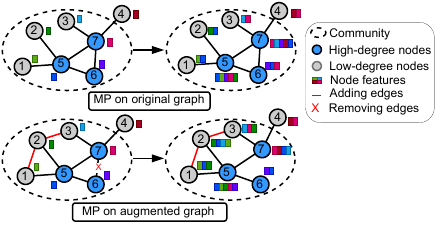}
    \caption{Message passing (MP) on original/augmented graphs.}
    \label{fig:problem_a}
\end{subfigure}
\hfill
\begin{subfigure}{0.47\textwidth}
\centering
    \includegraphics[width=\textwidth]{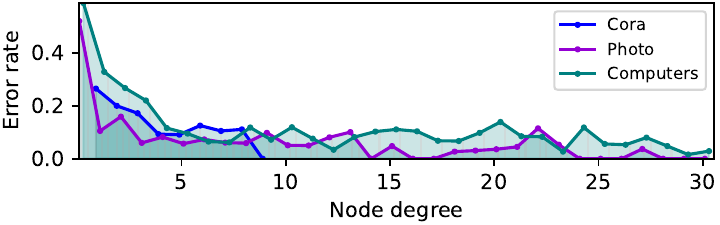}
    \caption{Misclassification rate of GT \cite{dwivedi2020generalization} method.}
    \label{fig:problem_b}
\end{subfigure}
\caption{ Message Passing GNNs is agnostic to the node degrees, resulting in low performance on low-degree nodes.
In (a), given an original graph (Above), low-degree nodes receive less information from the neighborhood, while high-degree nodes obtain too much information.
After adding within-community edges for low-degree nodes and removing edges connecting high-degree nodes by graph augmentation (Below), each node could receive adequate information.
In (b), low-degree nodes are more misclassified than high-degree nodes in the graph transformer model (GT \cite{dwivedi2020generalization}) on Cora, Photo, and Computers datasets.
}
\label{fig:problem_ab}
\end{figure}


One of the main problems in MP GNNs is that existing methods perform poorly on low-degree nodes, leading to degree bias in graph learning \cite{DBLP:conf/nips/Wang0SS22,DBLP:conf/cikm/LeeH0P22}.
The degree bias originates from low-degree nodes having only a few neighbors, while high-degree nodes have too many neighbors in most real-world graphs.
Furthermore, due to the power law distribution, most nodes in the graph have low degrees, while there are a few high-degree nodes.
This issue can negatively affect the MP GNNs' ability to learn low- and high-degree nodes.
First, low-degree nodes only receive a few messages from neighborhoods, which could be biased or even under-represented in graph learning \cite{d098f7454ebe43e082f5e94c7ff051d3}.
Second, high-degree nodes receive excessive information, which may lead to inherent limitations of GNNs, such as over-smoothing and over-squashing problems \cite{HUANG2023110556}.
However, most recent GNNs overlook degree biases, which could cause unfairness and poor performance in graph learning.
To illustrate the problem, consider four low-degree nodes ($1$, $2$, $3$, and $4$) in Figure \ref{fig:problem_ab}(a).
These nodes with few neighbors receive less information from the neighborhood compared with high-degree nodes (blue nodes). 
As shown in Figure \ref{fig:problem_ab}(b), the current graph transformer model, GT \cite{dwivedi2020generalization}, performs poorly on low-degree nodes on Cora, Photo, and Computers datasets.

Because of the lack of messages, most recent studies have proposed
augmentation-based strategies to provide more messages to low-degree nodes.
Several methods, i.e., RGRL \cite{DBLP:conf/cikm/LeeH0P22} and GRADE \cite{Zhu:2020vf}, combine the power of GCN and contrastive learning to learn representations via numerous views of the input graphs.
These methods commonly create multiple views of the input graph via heuristic or uniform random augmentations and then optimize a GNN encoder by contrasting between positive and negative samples.
However, the uniform random augmentations may not always generate valuable edges in different strategies \cite{DBLP:conf/iclr/LingJLJZ23}.
Additionally, we argue that edges connecting nodes within communities are more significant than others, which has not been fully explored.
For example, the uniform random augmentations may generate redundant edges, i.e., making fully connected graphs or removing too many edges, resulting in a loss of the graph structure.
Other studies aim to capture more messages for each node by expanding the range of the neighborhood for each target node \cite{dwivedi2020generalization,DBLP:conf/nips/KreuzerBHLT21,DBLP:journals/corr/abs-2308-09517}.
For example, SAT \cite{DBLP:conf/icml/ChenOB22} extract the $k$-subgraph information to update the vector representations for each target node.
However, these methods do not mainly solve the degree-unbiased problem \cite{lee2020story,DBLP:journals/sensors/JeonCL22,lee2021learning}.
Moreover, this could cause inherent limitations of GNNs, such as over-smoothing and over-squashing problems \cite{HUANG2023110556,nguyen2023companion,lee2021plot}.
Therefore, it is necessary to learn degree-unbiased representations with a single framework.

In this study, we propose a novel framework for learning degree-unbiased representations via learnable augmentations and graph transformers by extracting within-community structures.
The main challenge of designing learnable graph augmentations is how to provide valuable connection opportunities for nodes with the goal of obtaining node degree unbiases.
Thus, given a target node, it is important to capture ranked context nodes in order to generate the edges via learnable graph augmentations.
We then investigate the connections between nodes within the same community as they could share similar features.
Intuitively, we rank the context nodes within the community according to a degree score \textit{w.r.t} (i) they are in the same community and within a $k$-hop distance, and (ii) they have the same low degree together.
The graph augmentation is end-to-end trainable through edge perturbation, making itself learn the informativeness and degree unbiases.
To illustrate our augmentation strategy, consider three low-degree nodes $1$, $2$, and $3$ in Figure \ref{fig:problem_ab}(a).
The three nodes have similar low degrees and are connected to a dense topology within a community.
Therefore, our learnable augmentation module will construct edges between them with a high probability, which provides more within-community features.
The augmentation module can also remove edges between high-degree nodes via edge perturbation, i.e., the edge connecting node $6$ and $7$, shown in Figure \ref{fig:problem_ab}(a).

Note that the augmentation module could generate many edges connecting distant nodes even in the same community, which makes the model agnostic to learn the high-order proximity between pairs of nodes.
In addition, within a community, nodes with similar degrees (roles) tend to share similar features that also need to be discovered in graph learning.
Thus, we propose an improved self-attention, an extension of transformer attention to capture the high-order proximity and the node roles information between nodes.
It is worth noting that we directly encode the high-order proximity into full dot product attention, which could enable CGT to discover the proximity information along with messages from neighborhoods to target nodes.
Furthermore, the augmentation module could collapse into insignificant connections and thus could fail to generate appropriate graph data.
For example, the augmentation module could learn to generate a fully connected graph or remove too many edges that retain no original graph structure.
Such graph augmentations are not informative as they lose all the structural information from the original graph.
Therefore, we propose a self-supervised learning (SSL) task to preserve the $k$-step transition probability between node pairs and regularize the graph augmentations.
To the extent of our knowledge, CGT is the first graph transformer model addressing degree fairness.

In summary, our contributions are as follows:

\begin{itemize}
    \item We propose the utilization of within-community structures in learnable augmentations to allow low-degree nodes to be sampled with high probabilities via edge perturbation.
    \item  We propose a novel graph transformer with improved self-attention that takes the augmented graph as input and encodes the within-community proximity into dot product self-attention and the roles of nodes.
    \item We propose a self-supervised learning task to preserve graph connectivity and regularize the augmented graph data to generate the representations. 
\end{itemize}

\section{Related work}

Currently, several studies have proposed augmentation-based methods to achieve fair representations of low-degree nodes \cite{DBLP:conf/nips/YouCSCWS20,DBLP:conf/nips/Wang0SS22}.
These methods commonly construct multiple views via augmentations of the input graph and then optimize the representations through different contrastive learning strategies.
For example, GRACE \cite{Zhu:2020vf} first constructs two augmented views of a graph by randomly dropping edges or masking their features. 
Then, it aims to make positive representations to be close while pushing negative pairs far apart.
GCA \cite{DBLP:conf/www/0001XYLWW21} enhances GRACE by proposing adaptive augmentation techniques that focus on the graph structure and node features.
These methods use the probability as the weight of negative samples and treat all neighbor nodes as negative samples, which can not fully capture the global graph structure.
RGRL \cite{DBLP:conf/cikm/LeeH0P22} aims to preserve the global relationship between nodes with a heuristic augmentation for low-degree nodes.
GRADE \cite{DBLP:conf/nips/Wang0SS22} explores the context nodes of low-degree nodes within an ego network to obtain structural fairness.
Despite the success of the methods, the existing augmentation models that adopt heuristic augmentation operations or uniform random sampling, which are non-differentiable and thus could prevent the model from learning useful information.
Furthermore, the heuristic augmentations could fail to generate the appropriate graph data, such as dropping most of the edges or adding too many redundant edges.
In contrast, we use learnable augmentations to generate augmented graphs with degree-unbiased learning while preserving the most original graph structure.

Some recent approaches aim to expand the range and the sub-structures of neighborhoods of nodes to find more useful information.
For example, MixHop \cite{DBLP:conf/icml/Abu-El-HaijaPKA19} updates representations based on the neighbors within $k$-hop distance and then adjusts the aggregation mechanism. 
The SAT model \cite{DBLP:conf/icml/ChenOB22} learns representations by extracting multiple subgraphs at one time by replicating the nodes for every subgraph.
Several common graph transformers, i.e., GT \cite{dwivedi2020generalization} and SAN \cite{DBLP:conf/nips/KreuzerBHLT21}, treat graphs as fully connected graphs and enrich the high-order information for all nodes as adjacent connections.
Extending the $k$-hop neighborhood could mainly benefit low-degree nodes to obtain further information, as real-world graphs are dominated by low-degree nodes.
To the extent of our knowledge, there are no graph transformers targeting node degree unbiases.

\section{Methodology}
\begin{figure*}[t]
\centering 
  \includegraphics[width=1\linewidth]{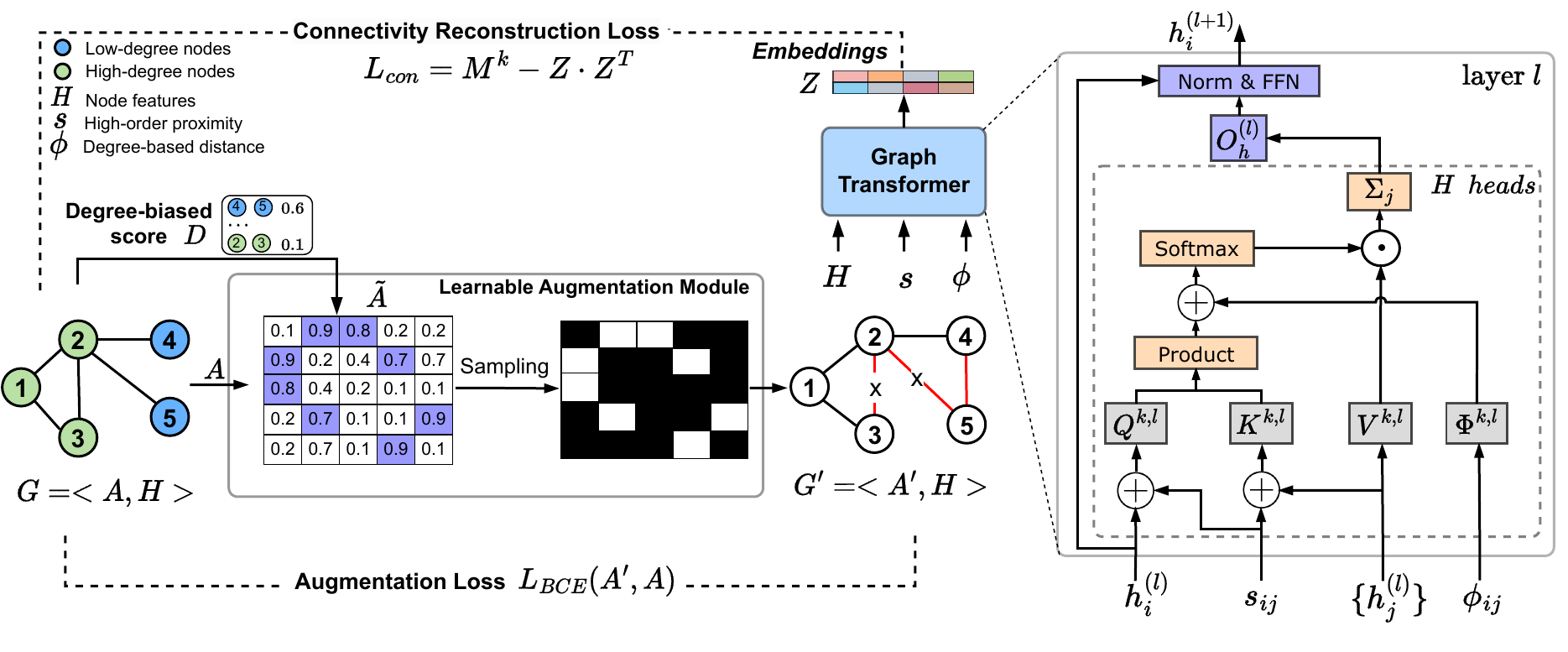}
  \caption{An overview of our framework.
  CGT comprises two main blocks, including learnable augmentation and graph transformer.
  }  
  \label{fig:model}
\end{figure*}

In this Section, we first introduce our strategy to build learnable graph augmentation (Sec. \ref{subsec:AutomatedGraphAugmentation}).
We then explain the CGT architecture in detail (Sec. \ref{subsec:graphtransformerarchitecture}).
Finally, we introduce the SSL task for graph structure preservation (Sec. \ref{subsec:ssl}). 
Figure \ref{fig:model} shows the overall architecture of our framework.

\subsection{Learnable Graph Augmentation}
\label{subsec:AutomatedGraphAugmentation}

CGT uses learnable augmentations to generate new graph data with degree unbiases by extracting community information while preserving the most original graph structure.
Given an input graph $G = (V, A)$ where $V$ denotes the set of nodes and $A$ is the adjacency matrix in $G$.
We then add to $A$ a matrix $D$, which contains ranked context nodes with a low-degree bias to generate a new matrix $\tilde{A}$.   
Thus, the matrix $\tilde{A}$ could contain useful information for providing opportunities for making connections between low-degree nodes within the same community.
We then generate a new graph $G' = (V, A')$ via edge transformation, as follows:
\begin{equation}
\tilde{A} = \xi A + \zeta   D \ , \tab  A' = \mathcal{T}_A ( \tilde{A} ) ,
\end{equation}
where $\xi$ and $\zeta$ are hyper-parameters to control the skewness of sampling. 
$\mathcal{T}_A$ is an edge perturbation transformation, which maps $\tilde{A}$ to the new adjacency matrix $A'$.
We now explain the strategies to construct the context nodes, ranked matrix $D$, and edge transformation.

\subsubsection{Capturing context nodes}
\label{subsubsection:Capturingglobalcontextnodes}

As mentioned before, we sample the nodes within the same community as they share similar features.
Given the graph $G$, we first cluster the nodes in the graph into $M$ partitions $G = \{G_1, G_2, \dots , G_M\}$ by applying the $K$-means clustering algorithm on the original graph to obtain $M$ clusters.
Intuitively, the context nodes are nodes that (i) can be reachable within $k$-step transition and (ii) are the same cluster as the target node.
We expect the context nodes to provide more valuable messages to the target nodes.

\begin{figure}[t]
\centering
  \includegraphics[width= .6\linewidth]{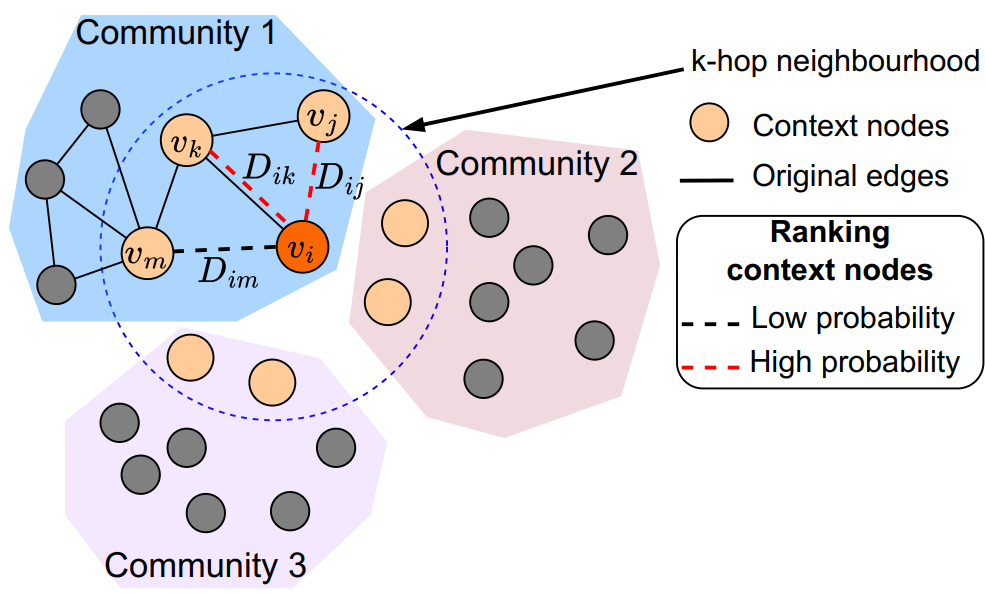}
  \caption{An overview of obtaining context nodes of the target node $v_i$.
  Given a target node $v_i$, we sample context nodes within the same community and then rank connections between $v_i$ and the context nodes as the probability for sampling based on Equation \ref{degree_bias}.}
  \label{fig:sampling}
\end{figure}

Formally, we define the set of context nodes of node $v_i$ as:

\begin{equation}
\label{neighbour_set}
    N(v_i) = \left\{ v_j \in V:A_{ij}^{(k)} > 0 \ , G_i = G_j \right\} , 
\end{equation}
where $A_{ij}^{(k)}$ denotes the $k$-step transition probability from $v_i$ to $v_j$.
$A_{ij}^{(k)} > 0 $ refers to the reachability between two nodes $v_i$ and $v_j$.
$G_i$ and $G_j$ are clusters of nodes $v_i$ and $v_j$, respectively.
By doing so, the connection between the target node $v_i$ and its context nodes will likely be sampled from our learnable augmentation.
To illustrate our sampling strategy, consider a target node $v_i$ and three clusters, as shown in Figure \ref{fig:sampling}.
The context nodes of $v_i$ are $v_j$, $v_k$, and $v_m$ as they are in the same community, and the distance between them is $k$-hop steps.

\subsubsection{Ranking context nodes}
\label{subsubsection:Capturingsimilaritywithlowdegreerelatedbiases}

The context nodes may contain a large set of high-degree nodes that could not be beneficial to making connections for low-degree nodes.
Thus, given a target node $v_i$, we now rank the set of context nodes $v_j$ with the goal of obtaining sampling for low-degree nodes with high probabilities.
Note that within a community, nodes with similar roles tend to share similar features, which could be beneficial to generating connections.
Therefore, the key idea is to apply a score from an inverse degree of nodes.
Precisely, we define the degree-biased score as:
\begin{equation}
\label{degree_bias}
   D_{ij}=\frac{1}{ \sqrt{d_i \cdot d_j}} , 
\end{equation}
where $d_i$ and $d_j$ denote the degree of nodes $v_i$ and $v_j$, respectively.
$D_{ij}$ refers to the row $i$-th and column $j$-th of the degree-bias matrix $D$.
By doing so, the connection between two nodes will be sampled with a high probability if they have a low degree together.

In a nutshell, given a target node $v_i$, we sample a set of context nodes $N(v_i)$ from Equation \ref{neighbour_set}, and then compute the degree-biased score towards low-degree nodes from Equation \ref{degree_bias}.
We then add the original adjacency matrix ($A$) to the degree bias matrix ($D$) to generate the $\tilde{A}$ matrix.
Afterward, the transformations for edges are performed through sampling from the Bernoulli distribution as described below.

\subsubsection{Edge perturbation}
\label{subsubsection:Edgeperturbation}

We now transform $\tilde{A}$ to a new adjacency matrix $A'$ through sampling from Bernoulli distribution parameterized with the probability in $\tilde{A}$, as follows:
\begin{equation}
\label{Bernoulli}
   A_{ij}^{'}=\text{Bernoulli}\left( \tilde {A}_{ij} \right) , \tab i, j = 1, \dots ,N.
\end{equation}

Note that the Bernoulli sampling function for the matrix $\tilde {A}$ is not differentiable.
Thus, to make the graph augmentation process differentiable in a fully end-to-end manner, we utilize the commonly-used scheme to approximate the Bernoulli sampling in Equation \ref{Bernoulli}.
Precisely, we approximate the Bernoulli sampling process by Gumbel-Softmax as a re-parameterization trick to relax the discrete distribution, following the work \cite{jang2017categorical,maddison2017the}.

\subsection{Community-aware Graph Transformer}
\label{subsec:graphtransformerarchitecture}

Given a graph $G = <V, E>$, the node feature $x_{i} \in R^{d_{0} \times 1}$ of node $v_{i}$ is projected via a linear transformation to $d$-dimensional hidden vector $h_{i}^{0}$, as $h_{i}^{0}={{W}_{0}}{{x}_{i}}+{{b}_{0}}$,
where ${W}_{0} \in R^{d \times d_{0} }$ and ${b}_{0} \in R^{d}$ are the learnable parameters of the linear transformation, $d_0$ is the original feature of $v_{i}$.

\subsubsection{Community-aware Self-Attention}
\label{sub:sa}

As mentioned above, the graph attention is agnostic to the high-order proximity and the roles of nodes within the community.
Thus, we aim to design an improved self-attention, an extension of transformer attention, which could learn the high-order proximity and node degree similarity between pairs of nodes.
We then inject the $QK$ vectors with high-order proximity between node pairs by concatenating the proximity $s$ and node feature $h$.
Accordingly, we use two separately learnable matrices for each weight matrix for learning node features and high-order proximity.
Precisely, we first project the high-order proximity vector to $d$-dimensional vectors, then add to the attention vectors of nodes.
The attention score at layer $l$-th of head $k$-th can be calculated, as follows:

\begin{align}
  & {{\alpha }_{ij}^{k,l}}=\frac{\left (Q^{k,l} \left[ h_{i}^{l}, s^{l}_{ij}\right ] \right)\left ({{K}^{k,l}} \left[ h_{j}^{l}, s^{l}_{ij}\right ] \right )}{\sqrt{{{d}_{k}}}}+{{\phi}_{f\left( i,j \right)}} , \\ 
 & Q^{k,l} \left[ h_{i}^{l}, s^{l}_{ij}\right ] =\left[ W_{n}^{Q^{k,l}}h_{i}^{l}+W_{s}^{Q^{k,l}} s^{l}_{ij} \right] ,\\ 
 & K^{k,l} \left[ h_{j}^{l}, s^{l}_{ij}\right ] =\left[ W_{n}^{K^{k,l}}h_{j}^{l}+W_{s}^{K^{k,l}}s^{l}_{ij} \right] ,
\end{align}
where ${W}^{Q^{k,l}}_n$, ${W}^{Q^{k,l}}_s$, ${W}^{K^{k,l}}_n$, ${W}^{K^{k,l}}_s$, ${\Phi}^{^{k,l}}$ $\in R^{d_{k} \times d}$, $k = 1$ to $H$ refers to the number of attention heads, $h_{i}$ and $h_{j}$ are the features of node $v_{i}$ and $v_{j}$, respectively.
$s_{ij}^{l}$ denotes the high-order proximity between two nodes $v_i$ and $v_j$, and $\phi_{f(i,j)}$ is a linearly transformed distance based on the role of nodes.

Our objective is to discover underlying within-community structures in graph learning. 
To address this situation, we propose a novel approach for calculating the high-order proximity between $v_i$ and $v_j$ in a multi-scaled manner as:

\begin{align}
  & {{s}_{ij}}=f\left( sim_{ij}^{(1)},sim_{ij}^{(2)},\ldots ,sim_{ij}^{(k)} \right) , \\ 
 & sim_{ij}^{(k)}=\frac{{{N}^{(k)}}(v_i)\cap {{N}^{(k)}}(v_j)}{{{N}^{(k)}}(v_i)\cup {{N}^{(k)}}(v_j)} \ ,
\end{align}
where ${{N}^{(k)}}(v_i)$ refers to the set of neighbours of $v_i$ up to $k$-step transition,
and $f(\cdot)$ denotes a linear transformation to $d$-dimensional hidden vectors.
To make CGT more sensitive to the roles of nodes with a low-degree bias, we introduce the function ${{\phi}_{f( i,j)}}$, which measures the distance between $v_i$ and $v_j$ based on their degree, as ${{\phi}_{f\left( i,j \right)}}=f\left( D_{ij} \right)$.
By utilizing  $\phi_{f(i,j)}$, each target node $v_i$ in the transformer layers can adaptively attend to other nodes according to the node roles.

\subsubsection{Graph Transformer Layers}
\label{Graph_Transformer_Layers}
We concatenate the outputs of the self-attention module into vector representations followed by a linear projection.
The node features $h^{l}_i$ of $v_i$ at layer $l$ is then updated as:
\begin{equation}
\label{eq:12}
\hat h_{i}^{l+1}=O_{h}^{l}\underset{k=1}{\overset{H}{\mathop{\mathbin\Big\Vert}}}\,\left( \sum\limits_{v_j\in N({{v}_{i}})}{\tilde{\alpha }_{ij}^{k,l}{{V}^{k,l}}h_{j}^{l}} \right),
\end{equation}
where ${\tilde{\alpha }_{ij}^{k,l}} = \text{softmax}_{j} ({{\alpha }_{ij}^{k,l}}  ) $, ${Q}^{k,l}$, ${K}^{k,l}$,${V}^{k,l}$ $\in R^{d_{k} \times d}$, ${O}^{l}_{h} \in  R^{d \times d}$, and $\mathbin\Vert$ refers to concatenation.
We then pass the outputs to feed-forward networks (FFN) by adding residual connections and layer normalization as:

\begin{align}
  & \hat{h}_{i}^{l+1}=\text{LN}\left( h_{i}^{l}+\hat{h}_{i}^{l+1} \right), \  
    h_{i}^{l+1}=W_{2}^{l} \sigma \left( W_{1}^{l}\hat{h}_{i}^{l+1} \right) 
\end{align}
where $W_{1}^{l} \in R^{2d \times d}$ and $W_{2}^{l} \in R^{d \times 2d}$ are learnable parameters, and LN is layer normalization.

\subsection{Self-Supervised Learning Tasks}
\label{subsec:ssl}

We provide an SSL task that could capture the graph connectivity and initial node features as well as regularize learnable augmentations.
To preserve graph connectivity and original node features, we aim to maximize the transition probability of all paths connecting pairs of nodes, following the work \cite{DBLP:journals/corr/abs-2308-09517,GutmannH12}, as:

\begin{align}
 L_1 = {\beta}_1 \sum\limits_{p}{   \left\| {M^{(p)} - Z^*} \right\|_{F}^{2}  }  + {\beta}_2 \frac{1}{\left| V \right|}{{{\left\| {X}-{{{\hat{X}}}} \right\|}_{2}}}  , 
\end{align}
where $M^{(p)}$ is the log scale matrix at $p$-th step transition matrix, $Z^*$ is the cosine similarity matrix calculated from representations, $X$ refers to original feature matrix, $\beta_1$ and $\beta_2$ are hyper-parameters.
$\hat{X}$ presents the learned representations after adding a linear layer to the representation $Z$.
Note that the augmentation module could collapse into insignificant connections and thus could fail to generate the appropriate graph data.
Thus, we use  a binary cross entropy (BCE) loss to penalize the graph augmentations, as follows:

\begin{align}
 \resizebox{0.42\textwidth}{!}{${{L}_{2}}=-\sum\limits_{i,j=1}^{N}{\left[ {{A}_{ij}}\log ( A_{ij}^{'} )+\left( 1-{{A}_{ij}} \right)\log ( 1-A_{ij}^{'} ) \right]}$}
\end{align}




The overall loss for the SSL task is then combined via a linear transformation, as follows:

\begin{align}
 L = {\alpha}_1 {L_{1}}  + {\alpha}_2 L_2, 
\end{align}
where ${\alpha}_1$ and $ {\alpha}_2$ are hyper-parameters.


Afterward, the learned representations are used to address downstream tasks.
In this study, we present two downstream tasks, including node classification and node clustering tasks.
For the node clustering task, we used modularity as the loss function, following the work \cite{TsitsulinPPM23}.
For the node classification task, we extracted the representations from one transformer layer to fully connected (FC) layers to obtain the classification output. 

\section{Evaluation}

In this Section, we first evaluate the performance of our proposed model against other state-of-the-art baselines on benchmark datasets.
We then examined the effectiveness of CGT and conducted ablation studies and sensitivity analysis. 


\subsection{Experimental Settings}

\subsubsection{Datasets}
We evaluate the performance of our proposed model by using various benchmark datasets on different downstream tasks.
Specifically, we used six publicly available datasets, which are grouped into three different domains, including citation network (Cora, Citeseer, and Pubmed datasets \cite{sen2008collective}), Co-purchase network networks (Amazon Computers and Photo datasets \cite{DBLP:conf/sigir/McAuleyTSH15}), and reference network (WikiCS \cite{DBLP:journals/corr/abs-2007-02901}).

\subsubsection{Baselines} 

We compare our proposed model to recent state-of-the-art GRL methods, including GNN variants, augmentation-based models, and graph transformers.
The GNN variants are GCN~\cite{DBLP:journals/corr/KipfW16}, 
GIN~\cite{DBLP:conf/iclr/XuHLJ19}, 
GAT~\cite{velivckovic2017graph},
and GraphSAGE~\cite{DBLP:conf/nips/HamiltonYL17}, 
We also compare our model with augmentation-based methods, such as RGRL \cite{DBLP:conf/cikm/LeeH0P22}, GRACE \cite{Zhu:2020vf}, GRADE \cite{DBLP:conf/nips/Wang0SS22}, and GCA \cite{DBLP:conf/www/0001XYLWW21}.
Furthermore, we compare CGT against recent graph transformers, such as GT~\cite{dwivedi2020generalization}, SAN~\cite{DBLP:conf/nips/KreuzerBHLT21}, SAT~\cite{DBLP:conf/icml/ChenOB22}, and GPS~\cite{DBLP:conf/nips/RampasekGDLWB22}.



\subsubsection{Implementation Details}
We conducted each experiment ten times by randomly sampling training, validation, and testing sets of size 60\%, 20\%, and 20\%, respectively.
The results written in the tables were measured with means and standard deviation on the testing set over the ten cases.
The hyper-parameters were tuned on the validation sets for each task and dataset.
For baselines, we follow the parameters obtained from the best variants. 

\subsection{Performance Analysis}


\subsubsection{Performance on Node Classification}

Table \ref{tab:node_classification} shows the performance of CGT compared to various baselines on the node classification task.
We have the following observations:
(i) Our proposed model performed well on all graph datasets compared with graph transformers that overlook the relations between low-degree nodes, i.e., GT, SAN, SAT, and GPS methods.
Since less information is delivered to low-degree nodes, our proposed model could be able to provide learning opportunities to low-degree nodes within communities, leading to the highest performance.
This demonstrates the benefit of our automated augmentation module and self-attention bias that could capture useful information between low-degree nodes.
(ii) CGT also outperformed augmentation-based contrastive learning methods, i.e., RGRL, GRACE, GRADE, and GCA models.
This is because our automated augmentation module is end-to-end trainable, leading to more effective learning representations on graphs.
In addition, by injecting within-community and node role similarity into self-attention, CGT could also learn the structure similarity between nodes.

\begin{table}[tb]
\centering
\begin{adjustbox}{width=.88\textwidth}
  \begin{tabular}{lcccccc}
    \toprule
    & Cora & Citeseer & Pubmed & WikiCS & Computers& Photo  \\\midrule \midrule
    GCN 
    & \textbf{85.89$\pm$1.06}
    & \textbf{73.20$\pm$1.08}
    & 85.74$\pm$0.61
    & 79.56$\pm$0.92   
    & 89.47$\pm$0.46
    & 93.38$\pm$0.50
     
    \\
    Sage 
    & \textcolor{orange}{\textbf{86.05$\pm$1.87}}
    & \textcolor{orange}{\textbf{74.58$\pm$1.33}}
    & 86.48$\pm$0.38
     & 82.90$\pm$0.80
    & 89.47$\pm$0.45
    & 92.28$\pm$0.58
   

    \\
    GIN 
    & 77.25$\pm$3.35
    & 64.09$\pm$1.95
    & 85.96$\pm$0.57
     & 76.53$\pm$0.82
    & 66.59$\pm$0.16
    & 88.92$\pm$1.36

    \\
    GAT 
    & 84.21$\pm$1.47
    & 73.43$\pm$1.21
    & 82.43$\pm$0.47
    & 77.55$\pm$0.71
    & 90.06$\pm$0.76
    & 93.34$\pm$0.73

    \\ 
   \midrule
    RGRL 
     & 84.27$\pm$0.87
     & 71.77$\pm$0.89
     & 82.50$\pm$0.17
     & 79.22$\pm$0.49
      & 84.83$\pm$0.43
      & 92.14$\pm$0.24
    
    \\
    GRACE 
    & 83.09$\pm$0.86
    & 69.28$\pm$0.29
    & 85.14$\pm$0.24
    & 30.52$\pm$0.54
    & 88.12$\pm$0.19
    & 92.21$\pm$0.10

    \\
    GCA 
    &83.43$\pm$0.34
    &68.20$\pm$0.26
    &85.65$\pm$1.02
    &32.34$\pm$0.07
    &74.87$\pm$0.11
    &91.28$\pm$0.58
    \\
    GRADE & 85.67$\pm$0.92
    &74.21$\pm$0.88
    &83.90$\pm$1.27
    &82.91$\pm$0.83
    &87.17$\pm$0.64
    &93.49$\pm$0.60\\
    
    \midrule
    
    GT 
    & 84.32$\pm$1.01
    & 72.51$\pm$1.65
    & \textcolor{orange}{\textbf{87.77}}$\pm$\textcolor{orange}{\textbf{0.60}}
    & \textbf{84.05$\pm$0.33}
    & \textbf{90.53$\pm$2.53}
    & \textcolor{orange}{\textbf{95.18}}$\pm$\textcolor{orange}{\textbf{0.66}}
    \\
    
    SAN 
    & 83.65$\pm$1.32
    & 72.12$\pm$1.89
    & 81.04$\pm$0.99
    & 81.04$\pm$0.81
    & 90.30$\pm$1.06
    & \textbf{95.08}$\pm$\textbf{0.48}
    \\
    SAT 
    & 79.13$\pm$0.73
    & 66.52$\pm$0.60
    & \textcolor{red}{\textbf{87.92}}$\pm$\textcolor{red}{\textbf{0.22}}
    & 80.04$\pm$0.76
    & 87.78$\pm$0.59  
    & 92.74$\pm$0.51
    \\
    GPS 
    & 75.64$\pm$0.94         
    & 65.71$\pm$0.59
    & OOM
    & 55.76$\pm$1.23
    & 50.26$\pm$5.95
    & 64.46$\pm$4.30
    \\
    \midrule
    Ours
    & \textcolor{red}{\textbf{87.10$\pm$1.53}}
    & \textcolor{red}{\textbf{76.59}}$\pm$\textcolor{red}{\textbf{0.98}}
    & \textbf{86.86}$\pm$\textbf{0.12}
    & \textcolor{red}{\textbf{84.61$\pm$0.53}}
    & \textcolor{red}{\textbf{91.45$\pm$0.58}}
    & \textcolor{red}{\textbf{95.73}}$\pm$\textcolor{red}{\textbf{0.84}}
     \\\bottomrule
\end{tabular}
\end{adjustbox}
\caption{The performance on node classification task (accuracy). 
  The highest three are highlighted by \textcolor{red}{\textbf{first}}, \textcolor{orange}{\textbf{second}}, and \textbf{third}}
\label{tab:node_classification}
\end{table}

\subsubsection{Performance on Node Clustering}
Despite the benefits of data augmentation, this can hinder the model from capturing the graph structure.
Therefore, we conducted further experiments on node clustering tasks to evaluate the performance of our model in understanding the graph structures.
For baseline models, we adopt the learned representations from node classification tasks and then use the $K$-mean clustering algorithm to partition nodes into clusters.
Tab \ref{tab:node_clustering} shows the performance on node clustering tasks of various methods.
We observed that:
(i) CGT could learn the graph structure well compared to baselines.
Although most GNN variants perform well on node classification tasks, they failed to capture the node partition and connectivity information.
It indicates that CGT relaxes the strict constraints of multiple views of the input graph, which can prevent the augmentation module from generating graphs that deviate too much from the input graph while capturing useful information from low-degree nodes.
(ii) Our model outperforms the augmentation-based methods, i.e., RGRL, GRACE, and GCA, as CGT could control the useful information from edge perturbations and the power of the self-attention bias.
This is because the augmentation module is end-to-en trainable, leading to the power to discover fairness-aware augmentations on degree-related biases, while current methods only use a heuristic augmentation.
Furthermore, the self-attention bias could enable our model to learn the within-community similarity between node pairs within the community.

\begin{table}[t]
\centering

\begin{adjustbox}{width=1\textwidth}
\begin{tabular}{l cc cc cc cc cc cc }
    \toprule
\multirow{1}{*}{} 
        & \multicolumn{2}{c}{Cora} & \multicolumn{2}{c}{Citeseer} & \multicolumn{2}{c}{Pubmed}  & \multicolumn{2}{c}{Computers} & \multicolumn{2}{c}{Photo} & \multicolumn{2}{c}{WikiCS} 
        \\
        \cmidrule(lr){2-3} \cmidrule(lr){4-5} 
        \cmidrule(lr){6-7} \cmidrule(lr){8-9}
        \cmidrule(lr){10-11} \cmidrule(lr){12-13}

         &C$\downarrow$& Q$\uparrow$
         &C$\downarrow$& Q$\uparrow$ 
         &C$\downarrow$& Q$\uparrow$ 
         &C$\downarrow$& Q$\uparrow$ 
         &C$\downarrow$& Q$\uparrow$
         &C$\downarrow$& Q$\uparrow$ 
         \\
    \midrule
     \midrule
\multirow{2}{*}{GCN }
    & 12.90 & \textcolor{red}{\textbf{70.53}}
    &  11.84    & 67.54
    & \textcolor{orange}{\textbf{9.54}}    & \textbf{54.59} 
    & 52.13   &  47.58
    & 21.57   & 69.82
    & 41.53  & 54.92
    \\
    &$\pm$0.26 & \textcolor{red}{\textbf{$\pm$0.28}}
    &$\pm$0.49    &$\pm$0.45
    & \textcolor{orange}{\textbf{$\pm$0.23}}    & \textbf{$\pm$0.12} 
    &$\pm$0.81    &$\pm$1.66
    &$\pm$3.34    &$\pm$4.60
    &$\pm$1.01  &$\pm$1.68
    \\

\multirow{2}{*}{Sage   }
    & 16.50&  66.69
    & 20.86& 59.21
    & \textbf{10.71}& 53.87
    & 23.29&74.29
    & 16.30& 79.77
    & 30.79& 67.22
    \\
    &$\pm$0.18&$\pm$0.26
    &$\pm$0.44&$\pm$0.46
    & \textbf{$\pm$0.44}&$\pm$0.34
    &$\pm$3.99&$\pm$3.65
    &$\pm$0.43&$\pm$2.37
    &$\pm$0.61&$\pm$1.05
\\

    \multirow{2}{*}{ GIN     }
   
     & 22.73&  59.52
    & 23.95&  55.35
     & 13.25&  49.01
     & 39.01&  59.57
     & 32.69&  62.53
     & 37.87&  60.48
    \\
     &$\pm$1.50&$\pm$1.93
    &$\pm$3.29&$\pm$3.38
     &$\pm$0.84&$\pm$0.95
     &$\pm$1.42&$\pm$1.19
     &$\pm$5.81&$\pm$6.69
     &$\pm$3.92&$\pm$3.90
    \\
\multirow{2}{*}{GAT     }
       & 16.05&{\textbf{ 67.29}}
    & 21.94& 58.17
    & 10.74& 53.21
    & 20.68& 77.41
     & \textbf{15.33}& \textbf{81.15}
      & 31.78& 66.38
    \\ 
       & $\pm$0.45&{\textbf{$\pm$0.35}}
    &$\pm$0.83&$\pm$0.69
    &$\pm$0.45&$\pm$0.76
    &$\pm$2.40&$\pm$2.08
     & \textbf{$\pm$0.42}& \textbf{$\pm$0.17}
      &$\pm$1.39&$\pm$2.06
    \\

    
    \midrule

    \multirow{2}{*}{RGRL }
     & \textbf{12.73}&  63.92
     & \textbf{5.66}& \textcolor{orange}{\textbf{ 68.41}}
      &$\pm$1.68& \textcolor{orange}{\textbf{ 56.36}}
     & \textcolor{orange}{\textbf{12.06}}&  \textcolor{orange}{\textbf{87.76}}
     & 17.57&  76.80
     & \textcolor{orange}{\textbf{22.62}}
&  \textcolor{red}{\textbf{77.47}} \\

     & \textbf{$\pm$1.44}&$\pm$4.96
     & \textbf{$\pm$0.37}& \textcolor{orange}{\textbf{$\pm$3.20}}
      &$\pm$1.68& \textcolor{orange}{\textbf{$\pm$15.39}}
     & \textcolor{orange}{\textbf{$\pm$1.28}}&  \textcolor{orange}{\textbf{$\pm$1.77}}
     &$\pm$3.62&$\pm$5.38
     & \textcolor{orange}{\textbf{$\pm$1.42}}
&  \textcolor{red}{\textbf{$\pm$2.41}}

     \\

     \multirow{2}{*}{GRACE     }

     & 22.22& 47.80
     & \textcolor{red}{\textbf{ 5.00}}& \textcolor{red}{\textbf{71.26}}
      & 14.13& 48.45
     & \textbf{13.56}&  85.09
     & \textcolor{red}{\textbf{9.32}}& \textcolor{orange}{\textbf{85.37}}
      & 51.97& 48.10
    \\
     &$\pm$0.06&$\pm$0.07
     & \textcolor{red}{\textbf{$\pm$0.04}}& \textcolor{red}{\textbf{$\pm$0.04}}
      &$\pm$0.04&$\pm$0.03
     & \textbf{$\pm$0.50}&$\pm$0.56
     & \textcolor{red}{\textbf{$\pm$0.00}}& \textcolor{orange}{\textbf{$\pm$0.00}}
      &$\pm$0.33&$\pm$1.15
    \\

      \multirow{2}{*}{GCA}

    &\textcolor{orange}{ \textbf{11.76}}    &   61.26
    & 13.52&   57.53
    & 20.80&   42.23
     & 13.69&   \textbf{86.21}
    & 21.73&   72.20
    & 38.95&   60.59
    \\
    &\textcolor{orange}{ \textbf{$\pm$2.05}}    &$\pm$2.64
    &$\pm$1.72&$\pm$1.50
    &$\pm$0.07&$\pm$0.09
     &$\pm$1.47&   \textbf{$\pm$2.08}
    &$\pm$5.63&$\pm$7.93
    &$\pm$6.05&$\pm$5.55
    \\

    \midrule
    \multirow{2}{*}{    GT  }

    & 17.77  & 65.07
    & 23.50  & 56.69
     & 19.16& 44.90
    & 26.53 & 72.77
    & 17.11& 78.47
    & 34.19 & 64.62
    \\
    &$\pm$0.81    &$\pm$0.83
    &$\pm$0.91  &$\pm$1.01
     &$\pm$0.99&$\pm$0.85
    &$\pm$6.77 &$\pm$7.68
    &$\pm$0.27 &$\pm$0.17
    &$\pm$3.57 &$\pm$4.18

    \\
        \multirow{2}{*}{      SAN}

    & 22.88   &  60.81
    & 24.48&  56.01
    &  14.89&  49.72
    & 30.61 &  67.36
    & 18.05 &  73.86
    & 30.22  &  67.95
    \\
    &$\pm$2.63     &$\pm$2.45
    &$\pm$1.95&$\pm$1.77
    &$\pm$1.20 &$\pm$1.08
    &$\pm$6.09 &$\pm$6.00
    &$\pm$1.71 &$\pm$2.63
    &$\pm$1.88  &$\pm$2.79
    \\

     \multirow{2}{*}{        SAT }

   & 28.25   &   54.09
    & 34.82&   45.31
     & 21.57&   43.04
     & 20.64&   79.07
    & 20.67&   71.85
    & 32.71&   66.96
    
    \\
   &$\pm$2.47    &$\pm$3.00
    &$\pm$2.94&$\pm$6.08
     &$\pm$1.97&$\pm$2.21
     &$\pm$5.15&$\pm$5.95
    &$\pm$7.44&$\pm$8.18
    &$\pm$3.84&$\pm$4.01
    \\
         \multirow{2}{*}{   GPS  }
 
     &  31.69   & 45.77
    &  44.06& 33.61
    &OOM &  OOM
    & 18.82&  81.28
     &  19.40&  75.11
     & \textbf{ 29.69}&  \textbf{68.56 }
    \\
     &$\pm$2.46    &$\pm$2.60
    &$\pm$4.62&$\pm$2.79
    &OOM &  OOM
    &$\pm$2.74&$\pm$2.93
     &$\pm$2.67&$\pm$3.87
     & \textbf{$\pm$1.56}&  \textbf{$\pm$2.90 }
\\


    \midrule
       \multirow{2}{*}{       Ours  }
       
     & \textcolor{red}{\textbf{ 9.84}}       &  \textcolor{orange}{\textbf{69.28}}
      &  \textcolor{orange}{ \textbf{ 5.40} }& \textbf{68.19}
    &  \textcolor{red}{\textbf{7.66}}&  \textcolor{red}{\textbf{ 89.51 }}
    & \textcolor{red}{\textbf{10.13}}&\textcolor{red}{\textbf{   88.07}} 
  & \textcolor{orange}{\textbf{9.71}} & \textcolor{red}{\textbf{ 85.39}}
   & \textcolor{red}{\textbf{21.68 }}&  \textcolor{orange}{\textbf{76.71 }}    
    \\
    
     & \textcolor{red}{\textbf{$\pm$0.76}}       &  \textcolor{orange}{\textbf{$\pm$0.32}}
      &  \textcolor{orange}{ \textbf{$\pm$1.21} }& \textbf{$\pm$0.39}
    &  \textcolor{red}{\textbf{$\pm$1.52}}&  \textcolor{red}{\textbf{$\pm$4.19 }}
    & \textcolor{red}{\textbf{$\pm$1.30}}&\textcolor{red}{\textbf{$\pm$1.32}} 
  & \textcolor{orange}{\textbf{$\pm$0.44}} & \textcolor{red}{\textbf{$\pm$2.58 }}
   & \textcolor{red}{\textbf{$\pm$0.82 }}&  \textcolor{orange}{\textbf{$\pm$1.31 }}    
    \\    
\bottomrule
\end{tabular}
\end{adjustbox}
\caption{The performance on node clustering task.
The top three are emphasized by \textcolor{red}{\textbf{first}}, \textcolor{orange}{\textbf{second}}, and \textbf{third}.}
\label{tab:node_clustering}
\end{table}

\subsubsection{Fairness Analysis}

\begin{figure}[tb]
\centering 
  \includegraphics[width= .81\linewidth]{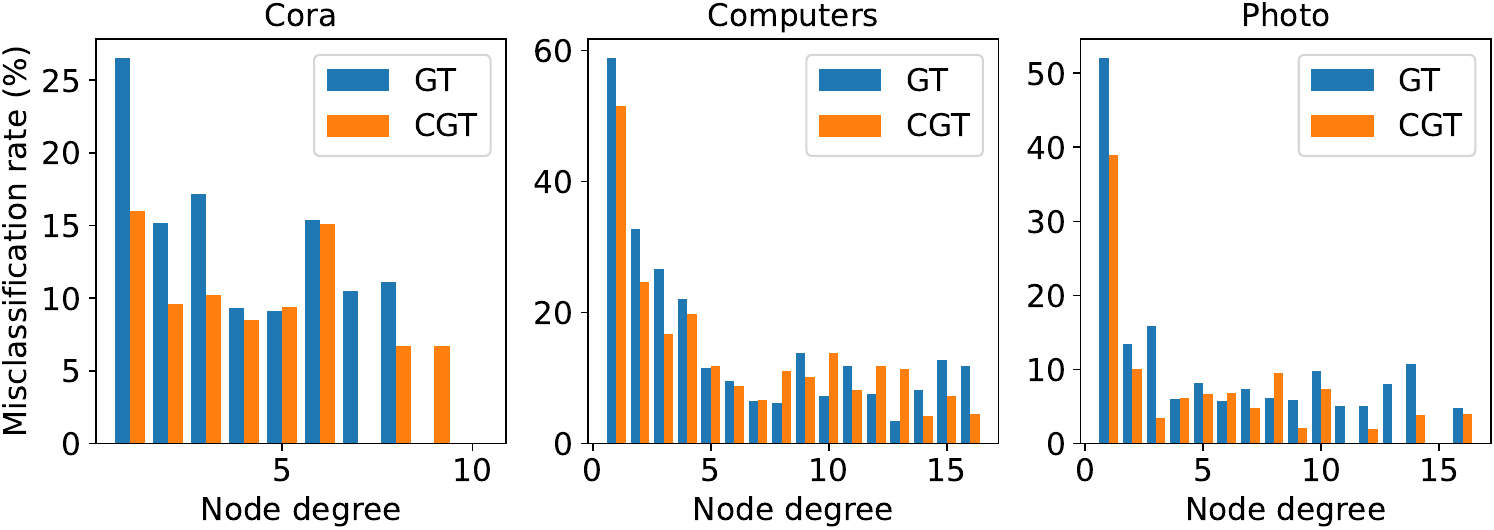}
  	\caption{Miss-classification rate on nodes w.r.t degrees between CGT and GT on Cora, Computers, and Photo datasets.} 
  \label{fig:Miss_classification}
\end{figure}

To validate the effectiveness of CGT on learning degree-related fairness, we further investigate the misclassification rate w.r.t node degrees, shown in Figure \ref{fig:Miss_classification}.
Furthermore, to deeply analyze the imbalance against baselines, we conducted the miss-classification rate on different ranges of low-degree node groups, as shown in Table \ref{tab:missclass_group}.
We observe that:
(i) When the node degree is very small, i.e., $ d \le 2$ and $ d \le 4$, the improvement of our model is more significant compared to the baselines. 
When the degree becomes larger, the improvement becomes smaller.
This is because when low-degree nodes have few neighbors, our augmentation module could likely generate more edges connecting low-degree nodes, thus significantly improving the performance on low-degree nodes.
In contrast, when the node degree goes higher, the models are already trained with adequate message information from neighborhoods, which makes the improvements minor.
(ii) CGT achieves the fairness of the miss-classification rate on different degree groups compared to baselines, especially on the Cora dataset.
The deviation of each node degree group stands relatively equal for nodes in different degree groups over the total error rate.
(iii) We observe that CGT significantly outperforms contrastive learning-based methods on low-degree nodes, i.e., RGRL, GRACE, and GCA. 
It indicates the superiority of our automated graph augmentation on the context nodes over the heuristic augmentations.

\begin{table}

\centering
\begin{adjustbox}{width=.85\textwidth}
\begin{tabular}{l cc cc cc cc cc c }
    \toprule

       \multirow{2}{*}{Method} & \multicolumn{2}{c}{ $ d \le 2$ } & \multicolumn{2}{c}{ $ d \le 4$ } & \multicolumn{2}{c}{ $ d \le 6$ }  & \multicolumn{2}{c}{ $ d \le 8$ } & \multicolumn{2}{c}{ $ d \le 10$ } & \multirow{2}{*}{All$\downarrow$} 
        \\
        \cmidrule(lr){2-3} \cmidrule(lr){4-5} 
        \cmidrule(lr){6-7} \cmidrule(lr){8-9}
        \cmidrule(lr){10-11} &
         Error$\downarrow$& Bias$\downarrow$
         &Error$\downarrow$& Bias$\downarrow$
         &Error$\downarrow$& Bias$\downarrow$
         &Error$\downarrow$& Bias$\downarrow$
         &Error$\downarrow$& Bias$\downarrow$\\
    \midrule
   
 
   \rowcolor{gray}   \multicolumn{12}{c}{ Cora } \\ 
 
    GIN &21.33&+4.61&18.69&+1.97&17.01&\textbf{+0.29}&17.12&+0.4&17.21&+0.49 &16.72 \\
 
    GAT&19.55&+6.4&\textbf{15.19}&+2.04&\textcolor{red}{\textbf{13.56}}&+0.41&\textcolor{red}{\textbf{13.71}}&+0.56&\textcolor{orange}{\textbf{13.58}}&+0.43& \textcolor{orange}{\textbf{13.51}}  \\\hline

  RGRL& 17.88&+2.95&15.82&+0.89&15.24&+0.31&15.15&+0.22&15.12&{\textbf{+0.19}}&14.92 \\
 GRACE &\textcolor{orange}{{\textbf{16.01}}}& {{\textbf{+1.59}}} &\textcolor{orange}{\textbf{15.02}}&+0.61&\textbf{14.7} &+0.28&\textbf{14.6}&\textbf{+0.19}&14.59&+0.17&  14.41 \\
  GCA & 19.19&+3.43&16.52&+0.76&15.81&\textcolor{orange}{\textbf{+0.05}}&16.85&+0.82&15.85&\textcolor{red}{\textbf{+0.09}}&15.76 \\ 
   GRADE& 17.94&+4.18&16.1&+2.34&15.42&+1.66&14.36&+0.60&\textbf{13.88}&+0.12&\textbf{13.76}\\\hline

  GT & 17.77& +2.01 &16.41 &+0.64&16.25 &+0.48&16.7 &+0.93&16.28 &+0.51 & 15.77 \\
 SAN & \textbf{16.09} & \textcolor{orange}{\textbf{+0.91}}&15.60 &\textbf{+0.41}&14.95 &\textcolor{red}{\textbf{-0.24}}&15.09 &\textcolor{orange}{\textbf{-0.11}}&15.37 &+0.18 & 15.19\\
    SAT&  26.34&+3.06&25.12&+1.84&24.38&+1.1&23.78&+0.5&23.71&+0.43& 23.28  \\
    Ours& \textcolor{red}{\textbf{12.51}} & \textcolor{red}{\textbf{-0.81}}& \textcolor{red}{\textbf{11.99}} & \textcolor{red}{\textbf{-1.32}}& \textcolor{orange}{\textbf{13.65}} & {+0.34}& \textcolor{red}{\textbf{13.05}} & \textcolor{red}{\textbf{-0.16 }}& \textcolor{red}{\textbf{13.47 }}& \textcolor{orange}{\textbf{+0.16}} & \textcolor{red}{\textbf{13.31}}  \\

 \hline 
  
     \rowcolor{gray}  \multicolumn{12}{c}{ Computers } \\ 
      GIN & 35.55&+22.54&25.96&+12.95&20.83&\textcolor{orange}{\textbf{+7.82}}&18.06&\textcolor{red}{\textbf{+5.05}}&\textbf{16.91}&\textcolor{red}{\textbf{+3.90}} & 13.01 \\
    
   GAT &36.93&+24.55&27.59&+15.21&\textbf{21.44} &+9.06&20.61&+8.23&18.21&+5.83& \textbf{12.38} \\\hline

      RGRL&\textcolor{orange}{\textbf{31.35}}&\textcolor{red}{\textbf{+16.71}}&27.17&\textbf{+12.53} &23.4&+8.76&21.65&+7.01&20.5&+5.86&14.63 \\
      
   GRACE &  33.37&+19.93&\textbf{25.91}&\textcolor{orange}{\textbf{+12.47}}&21.67&\textbf{+8.22}&19.32&\textbf{+5.87} &18.07&\textbf{+4.63}& 13.44\\
    
   GCA &37.83&\textbf{+17.9}&32.68&+12.75&28.77&+8.84&26.81&+6.88&25.95&+6.02& 19.93\\
    
  GRADE & 33.06&+20.61&28.83&+16.38&25.88&+13.43&23.46&+11.01&21.54&+9.09&12.45 \\
    
    \hline
    
  GT &37.02&+25.16&28.69&+16.83&24.5&+12.64&\textbf{21.09}&+9.23&19.92&+8.06& 11.86  \\
    
 SAN& 34.02&+24.61&\textcolor{orange}{\textbf{23.68}}&+14.27&\textcolor{orange}{\textbf{17.93}}&+8.52&\textcolor{orange}{\textbf{15.73}}&+6.32&\textcolor{orange}{\textbf{14.81}}&+5.39 & \textcolor{orange}{\textbf{9.41}} \\
    
    SAT& \textbf{32.42}&+19.16&27.43&+14.17&23.03&+9.77&20.95&+7.69&18.4&+5.14& 13.26  \\
     
   Ours& \textcolor{red}{\textbf{25.54}}& \textcolor{orange}{\textbf{+16.81}}& \textcolor{red}{\textbf{19.58}}& \textcolor{red}{\textbf{+10.85}}& \textcolor{red}{\textbf{15.68}}& \textcolor{red}{\textbf{+6.95}}& \textcolor{red}{\textbf{{13.91}}}& \textcolor{orange}{\textbf{+5.18}}& \textcolor{red}{\textbf{13.19}}& \textcolor{orange}{\textbf{+4.46 }}& \textcolor{red}{\textbf{ 8.73} } \\
    
        \hline
   
   \rowcolor{gray}       \multicolumn{12}{c}{ Photo } \\ 
    GIN & 38.61&+25.88&29.44&+16.71&30.29&+17.56&25.34&+12.61&24.67&+11.94& 12.73  \\
   
    GAT & 36.46&+26.8&28.72&+19.06&25.36&+15.7&20.16&+10.5&17.91&+8.25&9.66  \\

     \cmidrule(lr){1-12}
    
 RGRL& 21.91&+14.2&17.05&+9.35&15.16&+7.45&13.42&+5.72&12.82&+5.11&7.70\\
    
 GRACE & 21.5&+14.28&17.05&+9.84&14.86&+7.64&12.97&+5.76&12.22&+5.0 & 7.21 \\
    
  GCA & 21.7&+12.92&17.47&+8.69&16.27&+7.49&14.44&+5.66&14.15&+5.37&  8.77  \\ 
    
  GRADE & 18.06&+10.41&17.85&+10.21&15.46&+7.82&15.05&+7.41&14.01&+6.37& 7.64\\
    
    \hline
     
 GT& \textbf{14.78}&\textbf{+9.95}&\textbf{14.63}&+9.81&\textbf{12.98}&+8.15&\textbf{10.67}&+5.84&\textbf{10.5} &+5.67 & \textbf{4.83 } \\
    
 SAN & \textcolor{orange}{\textbf{13.25}}&\textcolor{orange}{\textbf{+8.58}}&\textcolor{orange}{\textbf{11.17}}&\textbf{+6.51}&\textcolor{orange}{\textbf{10.01}}&\textbf{+5.33}&\textcolor{orange}{\textbf{9.88}}&\textbf{+5.21}&\textcolor{orange}{\textbf{9.37}}&\textbf{+4.71}& \textcolor{red}{\textbf{4.67}}\\
    
SAT& 22.22&+9.53&16.04& \textcolor{red}{\textbf{+3.35}}&14.55&\textcolor{red}{\textbf{+1.86}}&13.46&\textcolor{red}{\textbf{+0.77}}&12.39&\textcolor{red}{\textbf{-0.3}}& 12.69 \\

    
 Ours&  \textcolor{red}{\textbf{11.83}}& \textcolor{red}{\textbf{+7.62}}& \textcolor{red}{\textbf{9.38}}& \textcolor{orange}{\textbf{+5.17}}& \textcolor{red}{\textbf{9.59}}&  {\textbf{+5.38}}& \textcolor{red}{\textbf{7.67}}& \textcolor{orange}{\textbf{+3.46}}& \textcolor{red}{\textbf{7.82}}& \textcolor{orange}{\textbf{+3.61 }}& \textcolor{red}{\textbf{ 4.21}}\\\hline
\end{tabular}
\end{adjustbox}
\caption{Miss-classification rate (\%) according to the range of node degree ($d$) on Cora, Computers, and Photo datasets in comparison between our proposed model and baselines.
For each degree group $d$, we measure the misclassification rate (Error) and the Bias compared to the overall error (All).
The top three are emphasized by \textcolor{red}{\textbf{first}}, \textcolor{orange}{\textbf{second}}, and \textbf{third}. 
}
\label{tab:missclass_group}
\end{table}

\subsection{Ablation Studies}

\begin{table}[tb]
\centering
 \begin{tabular}{c c cc  cccc }
    \toprule
    
      Pre.  &  Aug. &Att.& Cora & Computers &Photo   \\
     \hline \hline

\multirow{4}{*}{ - } &   -&  - 
& 83.91$\pm$1.29
&86.15$\pm$2.12  
&93.79$\pm$1.77\\ 

&   - &  $\checkmark$ 
&84.29$\pm$1.45
& 86.26$\pm$1.56
&94.54$\pm$1.87\\

&   $\checkmark$ &  - 
&84.06$\pm$1.17
&85.66$\pm$2.87
&94.67$\pm$1.52
  \\

&   $\checkmark$ &  $\checkmark$ 
&\textbf{85.36$\pm$1.21}
&87.57$\pm$1.79
&94.44$\pm$0.87
\\

\hline

\multirow{4}{*}{  $\checkmark$  } 
&  - &  - &
84.81$\pm$1.78
& \textbf{90.07$\pm$0.82}
&94.23$\pm$1.45\\ 

&   - &  $\checkmark$ 
&\textcolor{orange}{\textbf{85.54$\pm$1.07}}
&89.01$\pm$0.84 
&\textbf{95.08$\pm$1.08}\\

&   $\checkmark$ & - &
84.95$\pm$1.19 
&\textcolor{orange}{\textbf{90.38$\pm$0.41} }
& \textcolor{orange}{\textbf{95.46$\pm$0.23 }}
 \\

&   $\checkmark$ &  $\checkmark$ &
 \textcolor{red}{\textbf{87.10$\pm$1.53}}
& \textcolor{red}{\textbf{91.45$\pm$0.58}}
& \textcolor{red}{\textbf{95.73$\pm$0.84}}
  \\

    \hline
\end{tabular}
\caption{The effectiveness of different modules, including pre-training (Pre.), augmentation (Aug.), and self-attention (Att.).
The top three are emphasized by \textcolor{red}{\textbf{first}}, \textcolor{orange}{\textbf{second}}, and \textbf{third}. }
\label{tab:ablation_2}

\end{table}
To investigate the contribution of different components of CGT, we conducted further experiments on the Cora, Computers, and Photo datasets, shown in Table \ref{tab:ablation_2}.
We observe that:
(i) The use of pre-training leads to a significant improvement, especially in the Computer dataset.
It indicates that the augmentation module could not only benefit the low-degree nodes but also retain the original graph structures.
(ii) Another important contributing factor is the augmentation module on graphs, which provides connection opportunities for low-degree nodes.
It means the automated augmentation module delivers sufficient learning opportunities for low-degree nodes while reducing the bias of the transformers towards high-degree nodes.
In addition, attention modules contribute to a slight increase in the overall performance.

\subsection{Sensitivity Analysis}

\begin{figure}
	\centering
	\begin{subfigure}{.7\linewidth}
     \includegraphics[width= 1\linewidth]{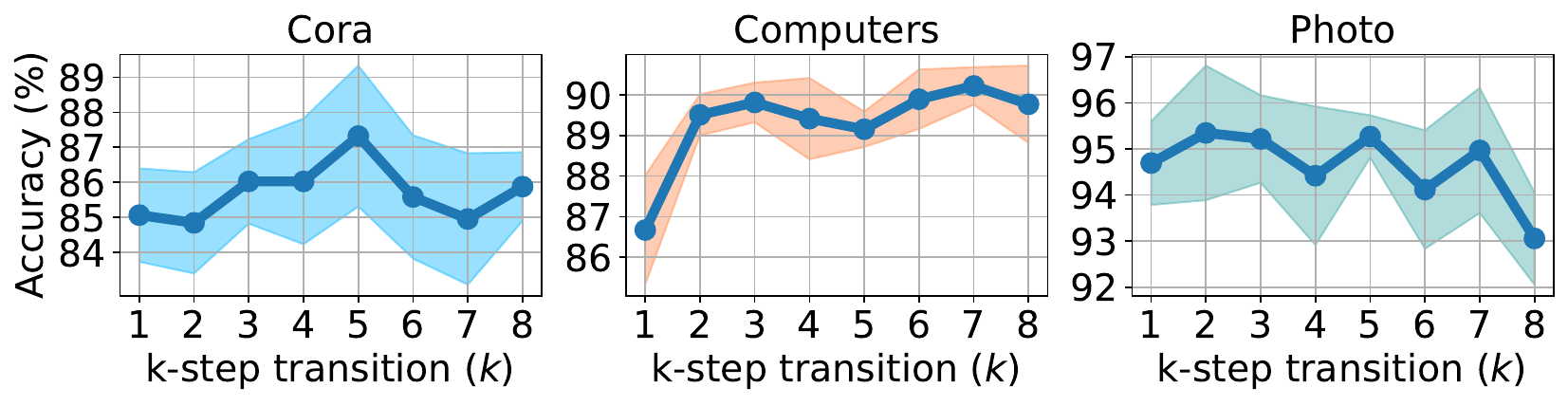}
            \caption{On the range of $k$-step transition.}
	\end{subfigure} 
	\begin{subfigure}{.7\linewidth}
     \includegraphics[width= 1\linewidth]{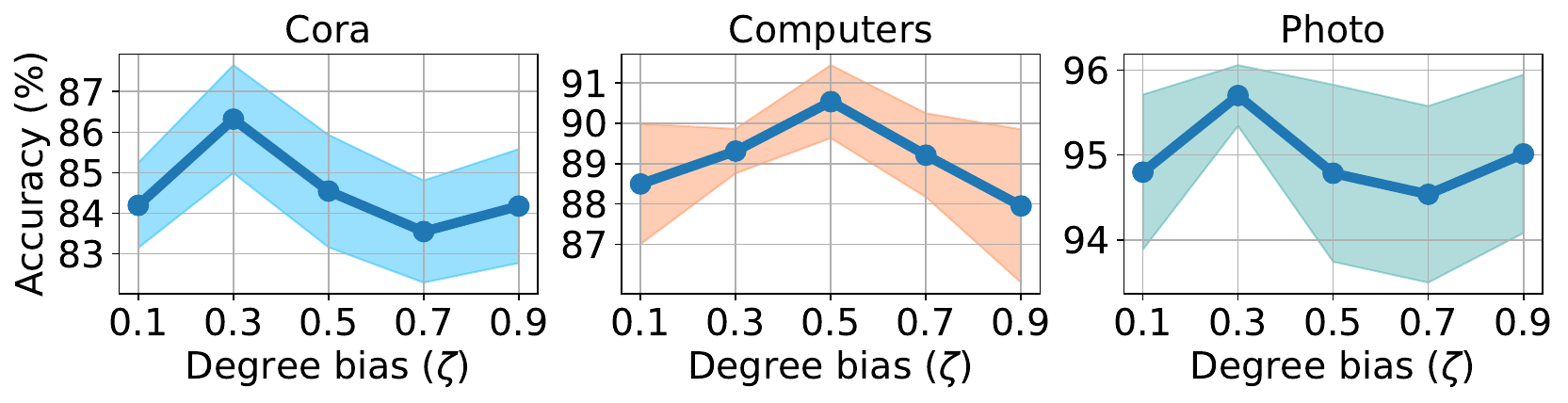}
            \caption{On the degree bias.}
	\end{subfigure} 
	\caption{The performance in terms of the sampling range of $k$-step transition (a) and the degree bias (b).}
	\label{fig:sas}
\end{figure}

In this section, we evaluate how the different options of the $k$-step transition and degree bias $d$ for the learned representations can affect the overall model performance, shown in Figure \ref{fig:sas}.
We observe that:
(i) The model performance increased significantly over the setting $k =2$ and $k=3$.
It indicates that the learned representations could learn not only through the adjacency connection but also can capture the global information from the neighborhood within communities.
(ii) When $k$ goes higher, the performance of the model reaches the maximum and then tends to decrease.
It means that the augmentation module could generate many unuseful edges, making it challenging for the model to learn the representations.
(iii) On the impacts of degree bias parameter $\zeta$, our model achieves the highest performance, i.e., $\zeta =0.3$ on the Cora and Photo datasets.
This is because the automated augmentation module could sample more low-degree nodes and make the adjacent connections between them, providing an opportunity for learning relations while reducing the connection between high-degree nodes.

\subsection{Complexity Analysis}

\begin{table}
\centering
  \begin{tabular}{lc}
    \toprule
    Model & Complexity \\\hline \midrule
    GT  & $O_{PE} (mE)$+ $O_{enc} (N^2)$\\
    SAN   & $O_{PE1} (mE)$ +  $O_{enc} (N^2)$ + $O_{PE2} (m^2 N)$\\ 
    SAT  &  $O_{PE} (mE)$ + $O_{subgraph} (N^k)$ + $O_{enc} (N^2)$ \\
    
    \hline
    CGT (ours) &  $O_{pre} (NE)$ + $O_{enc} (N^2)$ + $O_{bias} (N^2)$ \\\hline
\end{tabular}
\caption{Comparison on computational cost.}
\label{tab:complexity}
\end{table}

We compute three matrices, $d$, $s$, and $\phi$, only at the first time.
The computational cost of $d$ and $\phi$ for the whole graph is $O(N^2)$.
To construct $s$, we search on $k$-step matrices with the cost of $O(N^2)$.
For model steps, the computational cost for each layer in dot-product attention is $O(N^2)$, similar to other graph transformers.
SAT calculates representations in the time complexity of $O(N^k)$ to extract $k$-subgraphs for each representation.
In summary, CGT is more efficient than the SAN and SAT and only slightly increases complexity compared to the GT, as shown in Table \ref{tab:complexity}.

\section{Conclusion}

In this paper, we mitigate degree biases in the message-passing mechanism via learnable augmentation and graph transformers by extracting community structures.
Here, we propose learnable augmentations that could generate edges connecting low-degree nodes within communities with high probabilities. 
In addition, we propose improved self-attention, encoding the within-community and node role similarity between node pairs.
We also present in-depth discussions on how CGT achieves the best performance compared to recent methods.
Extensive experiments on real-world datasets demonstrate the effectiveness of CGT on downstream tasks as well as reducing the degree biases in the message-passing mechanism.
As mentioned above, CGT has a higher memory cost than other contrastive learning-based methods due to the quadratic complexity of full self-attention bias.
We plan to reduce the computational cost of the self-attention computation and extend CGT to the case that the community is sparse and has a limited number of nodes.

\section*{Acknowledgments}
This work was supported 
in part by the National Research Foundation of Korea (NRF) grant funded by the Korea government (MSIT) (No. 2022R1F1A1065516 and No. 2022K1A3A1A79089461) (O.-J.L.) 
and
in part by the Research Fund, 2022 of The Catholic University of Korea (M-2023-B0002-00088) (O.-J.L.). 

\bibliographystyle{unsrt}  
\bibliography{references}

\end{document}